\documentclass[10pt,twocolumn,letterpaper]{article}
\usepackage{amsmath}
\usepackage{algorithm}
\usepackage{booktabs}  
\usepackage{graphicx}  

\usepackage[camera-ready]{iccv}      
\usepackage{multirow}
\usepackage{booktabs}
\usepackage{graphicx}
\usepackage{multirow}

\usepackage{caption}  
\usepackage{algorithm}
\usepackage{algpseudocode}  

\definecolor{iccvblue}{rgb}{0.21,0.49,0.74}
\usepackage[pagebackref,breaklinks,colorlinks,allcolors=iccvblue]{hyperref}
\def\paperID{*****} 
\def\confName{ICCV}
\def\confYear{2025}

\title{RAIL: Region-Aware Instructive Learning for Semi-Supervised Tooth Segmentation in CBCT}

\author{
    \large Chuyu Zhao\textsuperscript{1\thanks{These authors contributed equally to this work.}}, 
    Hao Huang\textsuperscript{1\footnotemark[1]}, 
    Jiashuo Guo\textsuperscript{1\footnotemark[1]}, 
    Ziyu Shen\textsuperscript{1\footnotemark[1]}, 
    Zhongwei Zhou\textsuperscript{3\footnotemark[3]},
    Jie Liu\textsuperscript{1\thanks{Corresponding author: \href{mailto:jieliu@bjtu.edu.cn}{jieliu@bjtu.edu.cn}, \href{mailto:ykz@fudan.edu.cn}{yzk@fudan.edu.cn}}},
    Zekuan Yu\textsuperscript{2\footnotemark[2]} \\
    \normalsize \textsuperscript{1}School of Computer Science \& Technology, Beijing Jiaotong University, Beijing 100044, China \\
    \normalsize \textsuperscript{2}Academy for Engineering and Technology, Fudan University, Shanghai 200433, China \\
    \normalsize \textsuperscript{3}Department of Oral and Maxillofacial Surgery, General Hospital of Ningxia Medical University, Yinchuan 750004, China \\
    {\tt\small \{22723077, 22722088, 22722087, 23722061\}@bjtu.edu.cn, zzwjoel@hotmail.com}
}

\begin{document}
\maketitle
\begin{abstract}
Semi-supervised learning has become a compelling approach for 3D tooth segmentation from CBCT scans, where labeled data is minimal. However, existing methods still face two persistent challenges: limited corrective supervision in structurally ambiguous or mislabeled regions during supervised training and performance degradation caused by unreliable pseudo-labels on unlabeled data. To address these problems, we propose Region-Aware Instructive Learning (RAIL), a dual-group dual-student, semi-supervised framework. Each group contains two student models guided by a shared teacher network. By alternating training between the two groups, RAIL promotes intergroup knowledge transfer and collaborative region-aware instruction while reducing overfitting to the characteristics of any single model. Specifically, RAIL introduces two instructive mechanisms. Disagreement-Focused Supervision (DFS) Controller improves supervised learning by instructing predictions only within areas where student outputs diverge from both ground truth and the best student, thereby concentrating supervision on structurally ambiguous or mislabeled areas. In the unsupervised phase, Confidence-Aware Learning  (CAL) Modulator reinforces agreement in regions with high model certainty while reducing the effect of low-confidence predictions during training. This helps prevent our model from learning unstable patterns and improves the overall reliability of pseudo-labels. Extensive experiments on four CBCT tooth segmentation datasets show that RAIL surpasses state-of-the-art methods under limited annotation. Our code will be available at \href{https://github.com/Tournesol-Saturday/RAIL}{https://github.com/Tournesol-Saturday/RAIL}.
\end{abstract}

\section{Introduction}
\label{sec:intro}

Semi-supervised learning (SSL) has become a practical solution for 3D tooth segmentation in CBCT \cite{GAO20102406}, \cite{https://doi.org/10.1118/1.4901521}, \cite{JI2014116}, \cite{10.1007/978-3-642-35428-1_30}, \cite{Cui_2019_CVPR}, \cite{cui2022fully}, \cite{JING2024106032}, \cite{ZHONG2025108611}, where the cost and effort of manual annotation remain prohibitive in clinical-scale datasets. Semi-supervised methods \cite{cheplygina2019not}, \cite{tajbakhsh2020embracing}, \cite{yu2019uamt}, \cite{li2020sassnet}, \cite{luo2021dtc}, \cite{wu2021mcnet}, \cite{gao2024pmt}, \cite{chen2024ttmc}, \cite{bai2023bcp}, \cite{song2024SDCL} address annotation bottlenecks in medical imaging by exploiting a small labeled subset together with abundant unlabeled data.

Recent approaches in semi-supervised medical segmentation \cite{gao2024pmt}, \cite{song2024SDCL} employ primarily two key strategies: pseudo-labeling and consistency regularization. 

Pseudo-labeling \cite{cheplygina2019not}, \cite{tajbakhsh2020embracing}, \cite{yu2019uamt}, \cite{li2020sassnet}, \cite{luo2021dtc}, \cite{wu2021mcnet}, \cite{Cui_2019_CVPR} enables the model to generate provisional annotations for unlabeled inputs, which are subsequently leveraged as training signals. However, pseudo-labeling is prone to challenges, especially when the generated pseudo-labels are incorrect or unreliable \cite{cui2022fully}, \cite{JING2024106032}. Such inaccuracies can degrade the model’s performance, especially in regions with structural ambiguity caused by noisy data or complex anatomical features, where insufficient corrective supervision leads to inaccurate predictions.

In contrast, consistency-regularization~\cite{wang2023mcf}, \cite{JING2024106032} methods are designed to ensure that a model’s predictions remain stable for the same input across different perturbations, such as noisy data or random transformations. Recently, multimodel frameworks \cite{Chen_2021_CVPR}, \cite{luo2021dtc}, \cite{wang2023mcf}, \cite{GAO20102406}, \cite{Cui_2019_CVPR}, \cite{NEURIPS2023_7eeb4280}, \cite{gao2024pmt}, \cite{song2024SDCL} have been extensively applied to ensure stable and reliable predictions in medical image segmentation. However, these methods can lead to model imbalance, where inconsistencies in predictions across models can cause overfitting and the amplification of errors~\cite{ZHONG2025108611}.

To overcome such limitations, we propose the Region-Aware Instuctive Learning (RAIL), a dual-group, dual-student Mean Teacher framework for semi-supervised 3D CBCT segmentation, where the two groups are alternatively involved in training and gradient updating, allowing for a more balanced and effective model collaboration, leading to better generalization and reduced overfitting.

Specifically, we introduce two key modules: the Disagreement-Focused Supervision (DFS) Controller, which processes the differences between the student network output, ground truth, and the best student's output, guiding the model to focus on areas of structural ambiguity or incorrect labeling, and the Confidence-Aware Learning (CAL) Modulator that identifies regions of discrepancy between student network pseudo-labels and the best student pseudo-labels, ensuring the model places less emphasis on uncertain areas and reduces the impact of low-confidence predictions in unsupervised learning.

Our major contributions are summarized as follows: 
\begin{itemize}
    \item We propose a dual-group, dual-student Mean Teacher framework for semi-supervised 3D tooth segmentation from CBCT.
    \item We design a Disagreement-Focused Supervision (DFS) Controller to target areas with structural ambiguity or incorrect labeling.
    \item We design a Confidence-Aware Learning (CAL) Modulator to enhance pseudo-label reliability. 
\end{itemize}

Extensive experiments were conducted on four CBCT tooth segmentation datasets (FDDI+, FDDI-E, 3D CBCT Tooth, and CTooth) to evaluate the RAIL algorithm. The results demonstrate that RAIL achieves competitive performance under sparse supervision, outperforming prior methods with limited labeled data.

\captionsetup[figure]{labelfont=bf, labelsep=period, name=Fig.}  

\begin{figure*}[t]
  \centering
  \includegraphics[width=1\textwidth]{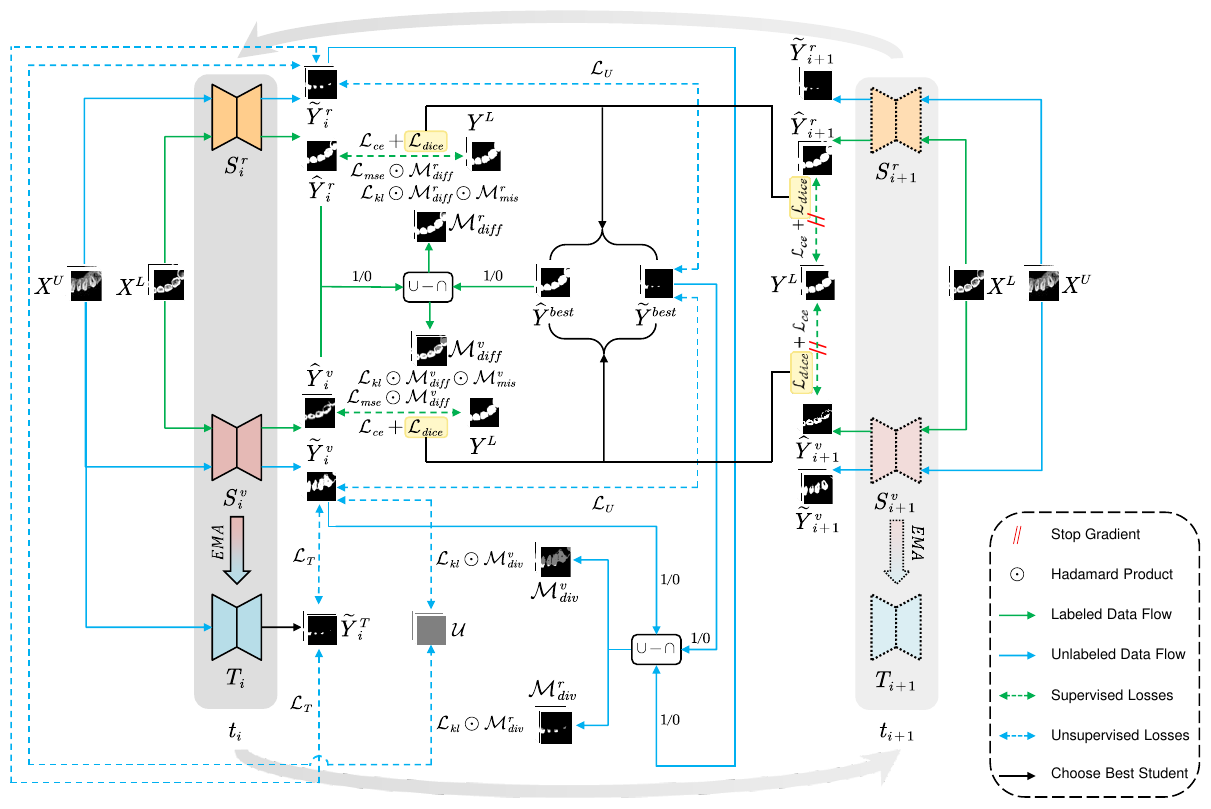}
  \caption{Pipeline of our Region-Aware Instructive Learning (RAIL) framework in Mean Teacher architecture. The total loss function for every student network in the training phase includes supervised losses $\mathcal{L}_s$, $\mathcal{L}_{DFS}$, and unsupervised losses $\mathcal{L}_U$, $\mathcal{L}_T$, $\mathcal{L}_{CAL}$.}
    \label{fig:pipeline}
\end{figure*}
\section{Related Work}
\label{sec:rela}

\subsection{Tooth segmentation in CBCT}

Tooth boundary extraction from CBCT images remains a persistent challenge due to anatomical complexity and imaging artifacts. Over the years, the field has witnessed a methodological shift—from classical algorithmic solutions to modern deep learning paradigms—reflecting significant progress in both accuracy and automation. Early methods, such as level-set and graph-cut algorithms, have laid the foundation for tooth segmentation. For instance, Gao et al. \cite{GAO20102406} applied level-set models enhanced with prior knowledge of shape and intensity distributions. Building on this, Gan et al. \cite{https://doi.org/10.1118/1.4901521} proposed a hybrid strategy that integrates multiple energy functionals to enable more accurate contour evolution during segmentation. Similarly, Ji et al. \cite{JI2014116} introduced a specialized level-set approach tailored for the segmentation of anterior teeth in CBCT scans. In addition, Graph-cut techniques have also been widely adopted, with Keustermans et al. \cite{10.1007/978-3-642-35428-1_30} incorporating statistical shape models to improve segmentation robustness. While effective under controlled conditions, these methods often depend on manual initialization and degrade under noise or anatomical ambiguity, limiting their accuracy and generalizability.

Deep learning has recently enhanced CBCT-based tooth analysis. Cui et al. \cite{Cui_2019_CVPR} introduced ToothNet, an end-to-end model for instance-level segmentation and classification, yielding superior accuracy compared to conventional approaches. Introduced an end-to-end artificial intelligence solution aimed at segmenting dental and alveolar structures from CBCT data, demonstrating resilience even in anatomically complex or artifact-prone scans. In a related advancement, Jing et al. \cite{JING2024106032} proposed a dual-phase semi-supervised framework that incorporates an Adaptive Channel Interaction Module (ACIM) alongside an uncertainty-guided regularization mechanism. In 2024, Hao et al. developed the T-Mamba architecture, which integrates a Tim block with DenseVNet to jointly leverage shared positional encodings and frequency-oriented representations. Zhong et al.~\cite{ZHONG2025108611} proposed a lightweight segmentation architecture named PMFSNet, which integrates a PMFS block to achieve an effective compromise between computational cost and segmentation precision in the context of dental imaging. However, deep learning methods often require extensive manual annotations and face challenges in handling limited annotated data. Accordingly, designing resilient architectures capable of integrating both global contextual cues and fine-grained local features remains essential, particularly in light of the scarcity of annotated samples and the anatomical intricacy inherent in dental structures. Our work addresses this by proposing Region-Aware Instructive Learning (RAIL), a novel framework that integrates dual-student models and employs region-aware instruction to improve segmentation under limited annotation.

\subsection{Semi-supervised learning in segmentation}

Semi-supervised learning (SSL) offers an effective solution to annotation scarcity in medical image segmentation \cite{cheplygina2019not}, \cite{tajbakhsh2020embracing}, including Ultrasound Computed Tomography (USCT) \cite{duric2007detection}. A common SSL strategy, pseudo-labeling \cite{iscen2019label}, suffers from low-confidence predictions due to insufficient labeled data. To mitigate this, consistency regularization methods enforce prediction consistency under perturbations, thereby enhancing model robustness.

Yu et al. \cite{yu2019uamt} introduced a self-ensembling strategy called UA-MT. The method leverages Monte Carlo dropout to estimate predictive uncertainty and reduce the influence of unreliable regions. Building on this idea, Li et al. \cite{li2020sassnet} developed SASSNet, which incorporates structural priors into a semi-supervised 3D segmentation framework to enhance anatomical fidelity. Further extending this line of work, Luo et al. \cite{luo2021dtc} proposed DTC, a dual-task architecture that concurrently predicts voxel-wise masks and geometric descriptors to preserve shape consistency during training. By enforcing consistency between these tasks, this framework significantly enhances segmentation accuracy while reducing the reliance on labeled data. Concurrently, Wu et al. \cite{wu2021mcnet} devised MC-Net, a mutual consistency-based training strategy for segmenting the left atrium, where predictive alignment across multi-view inputs is enforced to enhance segmentation reliability.

Building upon these early efforts, Gao et al. \cite{gao2024pmt} introduced a progressive mean teacher (PMT) framework that explores temporal consistency to improve segmentation accuracy over time. This approach, which builds on the Mean Teacher framework, employs exponential moving averages of model weights to guide the student network. In a further advancement of consistency-driven learning, Chen et al.\cite{chen2024ttmc} advanced consistency learning by unifying three tasks in TTMC for improved 3D analysis. Bai et al.\cite{bai2023bcp} later introduced BCP, a data augmentation technique that mixes labeled and unlabeled volumes bidirectionally to boost diversity under semi-supervised settings.

In addition, shape-driven and discrepancy-aware methods have emerged to counteract prediction noise and pseudo-label uncertainty. In this regard, Song and Wang \cite{song2024SDCL} introduced a student discrepancy-informed correction learning (SDCL) framework, which corrects pseudo-labels based on discrepancies between student models. 

Despite these advancements, SSL frameworks still face challenges in medical imaging tasks, particularly for handling variability in image quality, anatomical complexity, and the need for more reliable pseudo-labeling methods. To address these challenges, our work introduces two key contributions: a Confidence-Aware Learning Modulator (CAL) that enhances pseudo-label reliability by focusing on high-confidence regions and minimizing the impact of low-confidence areas, and a Disagreement-Focused Supervision (DFS) Controller, which targets regions where model predictions diverge. These mechanisms improve pseudo-label reliability, model stability, and segmentation accuracy, particularly in anatomically complex or ambiguous areas, thus enhancing performance under limited annotations.

\subsection{Multimodel Framework}

Multimodel architectures have emerged as a pivotal strategy in semi-supervised learning (SSL), especially within the domain of medical image segmentation. These frameworks capitalize on the use of multiple networks or their variants to enforce output consistency, thereby enhancing both model robustness and generalization capability. For instance, Chen et al. \cite{Chen_2021_CVPR} proposed Cross-Pseudo Supervision (CPS), where two networks iteratively exchange pseudo-labels to enable collaborative training. Moreover, the dual-task consistency framework, introduced by Luo et al. \cite{luo2021dtc}, enforces consistency across multiple tasks within a single model, indicating the effectiveness of multitask learning in semi-supervised settings. 

Currently, the use of heterogeneous models has been further explored to refine output consistency. For instance, Wang et al. \cite{wang2023mcf} presented a Mutual Correction Framework (MCF), which employs heterogeneous models to constrain output consistency, thereby enhancing segmentation accuracy under semi-supervised settings. This approach underscores the benefits of model-level regularization in multimodel frameworks. Additionally, Na et al. \cite{NEURIPS2023_7eeb4280} introduced a multiteacher framework, where multiple teachers are used to promote diverse learning in semi-supervised semantic segmentation. It highlights the importance of maintaining diversity among models to prevent overfitting and strengthen generalization.

However, multimodel frameworks can suffer from inefficiencies due to overfitting or the propagation of incorrect predictions across models. Our work introduces a dual-group, dual-student framework where inter-group knowledge transfer is promoted, allowing for a more balanced and effective model collaboration, leading to better generalization and reduced overfitting.
\section{Methodology}
\label{sec:method}

\subsection{The Overall Pipeline of RAIL}

\subsubsection{Problem Definition}

Our training medical image dataset $\mathcal{D} = \{\mathcal{D}^l, \mathcal{D}^u\}$ contains $N$ labeled images $\mathcal{D}^l$ and $M$ unlabeled images $\mathcal{D}^u$ ($N \ll M$), where $\mathcal{D}^l = \{(x_i^l, y_i^l)\}_{i=1}^N$ and $\mathcal{D}^u = \{x_i^u\}_{i=N}^{N+M}$. Each 3D volume image $x_i^l \in \mathbb{R}^{W \times H \times D}$ in $\mathcal{D}^l$ has a label $y_i^l \in \{0, 1\}^{W \times H \times D}$, where 0 denotes the background class and 1 represents the foreground target. The model produces an output prediction denoted as $\hat{y}_i \in \{0, 1\}^{W \times H \times D}$, representing a volumetric segmentation across spatial dimensions. Each time a batch is fed into the network, it contains an equal proportional volume of labeled data $(X^L, Y^L)$ and unlabeled data $X^U$. Predictions generated by the network for labeled samples are represented as \( \widehat{Y} \), while those corresponding to unlabeled inputs are denoted by \(\widetilde{Y} \). 

We have incorporated the Mean Teacher architecture within the semi-supervised learning framework. This integration aims to enhance the model by providing high-quality pseudo-labels while also ensuring that the model's structure facilitates continuous improvement in its representational power. This leads to enhanced performance while maintaining robust diversity. The teacher network mirrors the student network in structure but plays a passive role in training. Parameter updates are performed via the Exponential Moving Average (EMA) mechanism, which enforces consistency regularization on the student network by leveraging its predictions. The update follows the EMA formulation given below:
\begin{equation}
\theta'_t = \alpha \theta'_{t-1} + (1 - \alpha) \theta_t
\label{ema_formula}
\end{equation}
Here, $\theta'_t$ represents the teacher model’s parameters at the current iteration, while $\theta'_{t-1}$ corresponds to its parameters from the previous step. The student model’s parameters at the current iteration are denoted by $\theta_t$. The hyperparameter $\alpha$ serves as a momentum-like smoothing factor that governs the update rate of the teacher network.

\subsubsection{Dual-group Dual-student Mean Teacher Framework}

The training framework of RAIL is shown in Fig.~\ref{fig:pipeline}. RAIL consists of two groups of Mean Teacher networks with the same framework, which are alternatively involved in training iterations. $t_i$ denotes the current training iteration and $t_{i+1}$ denotes the next turn of the training iteration. Each group of frameworks contains two student networks and one teacher network, where $S^v$ denotes the VNet student network, $S^r$ denotes the ResVNet~\cite{wang2023mcf} student network, and $T$ denotes the VNet teacher network. Following established practices, we employ the VNet-based student model $S^v$ to update the teacher network $T$ via the Exponential Moving Average (EMA) scheme.

The training process of RAIL consists of three parts: (i) obtaining some fundamental supervised and unsupervised losses according to the PMT strategy; (ii) Disagreement-Focused Supervision (DFS) Controller: generating a DiffMask $\mathcal{M}_{diff}$ from the difference between student segmentation and the best student segmentation, and a MisMask $\mathcal{M}_{mis}$ from the misalignment between student segmentation and ground truth, thus multiplying $\mathcal{M}_{diff}$ and $\mathcal{M}_{mis}$ to create a DiffMisMask $\mathcal{M}_{diffmis}$ to guide the model in focusing supervision on structurally ambiguous or mislabeled voxels; (iii) Confidence-Aware Learning (CAL) Modulator: generating DivMask $\mathcal{M}_{div}$ from the divergence between the student pseudo-label and the best student pseudo-label to improve the overall stability and reliability of the pseudo-label. For convenience, the upper and lower corner notations of many symbols are simplified here. A more detailed symbolic description is given in later explanations.

\subsection{Progressive Mean Teacher}

\begin{algorithm}[htbp]
\caption{Training with Confidence-Aware Learning}
\label{CAL Algorithm}
\textbf{Input:} Student networks $S^v_i$, $S^r_i$, $S^v_{i+1}$, $S^r_{i+1}$; teacher network $T_i$; labeled dataset $\mathcal{D}^l = \{(x_i^l, y_i^l)\}_{i=1}^N$; unlabeled dataset $\mathcal{D}^u = \{x_i^u\}_{i=N}^{N+M}$; boolean flag $first\_term = True$ for initial phase; number of classes $K = 2$. \\ 
\textbf{Output:} \parbox[t]{\linewidth}{Updated weights for $S^v_i$, $S^r_i$.} \vspace{-0.47cm}
\begin{algorithmic}[1]
\For{each training iteration $t_i$}
    \State Sample a batch $(X^L, Y^L)$, $X^U$ from $\mathcal{D}^l$ and $\mathcal{D}^u$
    \Statex \texttt{// Supervised training without CAL}
    \If{$first\_term$}
        \For{$S \in \{S^v_i, S^r_i\}$}
            \State Update $S$ by backpropagating $\mathcal{L}_{DFS}$
        \EndFor
        \State Update $T_i \leftarrow \mathrm{EMA}(S^v_i)$
        \State \textbf{continue} to next iteration
    \Statex \texttt{// Semi-supervised training with pseudo-labels and CAL algorithm}
    \Else
        \Statex \texttt{// Compute supervised Dice loss on labeled data for all students}
        \For{$S \in \{S^v_i, S^r_i, S^v_{i+1}, S^r_{i+1}\}$}
            \State $\mathcal{L}_{dice}^S = DiceLoss(S(X^L), Y^L)$
        \EndFor
        \Statex \texttt{// Identify the best student and get its pseudo-label}
        \State $S^{best} \gets \arg\min_S \mathcal{L}_{dice}^S$
        \State $\widetilde{Y}^{\text{best}} = S^{best}(X^U)$
        \For{$S \in \{S^v_i, S^r_i\}$}
            \State $\widetilde{Y}^{v/r}_i \gets S^{v/r}_i(X^U)$
            \Statex \texttt{// Compute $\mathcal{M}^{v/r}_{div}$: pixels where $S^{v/r}_i$ disagrees with $S^{best}$}
            \State $\mathcal{M}^{v/r}_{div} = \widetilde{Y}^{v/r}_i \oplus \widetilde{Y}^{\text{best}}$
            \Statex \texttt{// uniform distribution tensor}
            \State $\mathcal{U}(k) \gets 1/K$, $\mathcal{U} = [1/2, ..., 1/2]$
            \State $\mathcal{L}^{v/r}_{CAL} = \mathcal{D}_{KL}(\widetilde{Y}^{v/r}_i \| \mathcal{U}) \odot \mathcal{M}_{div}$
            \State Update $S$ by $\mathcal{L}_{CAL} + \mathcal{L}_{DFS}$
        \EndFor
        \State Update $T_i \gets \text{EMA}(S^v_i)$
        \State \textbf{continue} to next iteration
    \EndIf
\EndFor
\end{algorithmic}
\end{algorithm}

Our framework is integrated with the state-of-the-art PMT method~\cite{gao2024pmt} to enhance performance. The supervised loss, denoted as $\mathcal{L}_s$, is defined as follows:
\begin{equation}
\mathcal{L}^{v/r}_s = \mathcal{L}_{CE}(\widehat{Y}^{v/r}_i, Y^L) + \beta \mathcal{L}_{MSE}(\widehat{Y}^{v/r}_i, Y^L) \odot \mathcal{M}^{v/r}_{diff}
\end{equation}
where $\beta = 0.5$. $\widehat{Y}^{v/r}_i$ represents the model outputs for labeled data of the two student networks in the current training phase, and $Y^L$ denotes the ground truth. $\mathcal{L}_U$ represents the unsupervised loss:
\begin{equation}
\mathcal{L}^{v/r}_U = \mathcal{L}_{MSE}(\widetilde{Y}^{v/r}_i, \widetilde{Y}^{best})
\end{equation}

Additionally, the consistency loss $\mathcal{L}_T$, derived from the Mean Teacher framework, is computed as the mean squared error between the pseudo-labels $\widetilde{Y}^{v/r}_i$ and $\widetilde{Y}^T_i$ produced by the student and teacher networks:
\begin{equation}
\mathcal{L}^{v/r}_T = \mathcal{L}_{MSE}(\widetilde{Y}^{v/r}_i, \widetilde{Y}^T_i)
\end{equation}

\subsection{Disagreement-Focused Supervision}

In the training phase, we introduce the Disagreement-Focused Supervision (DFS) Controller, which minimizes the Kullback-Leibler Divergence~\cite{belharbi2021deep} between the student model's predictions and the ground truth. This approach encourages the model to focus its learning on regions where predictions are correct and where structural clarity is achieved, thereby enhancing the efficacy of supervised learning.

First, the model outputs of the two student networks $\widehat{Y}^v_i$, $\widehat{Y}^r_i$ in the current training phase, and the two student networks $\widehat{Y}^v_{i+1}$, $\widehat{Y}^r_{i+1}$ in the next turn of the training phase, perform a DICE loss calculation with ground truth $Y^L$, respectively ($\widehat{Y}^v_{i+1}$ and $\widehat{Y}^r_{i+1}$ do not perform gradient updating). We choose the student network with the highest DICE loss as the current best student, whose corresponding label predictions and pseudo-labels are denoted as $\widehat{Y}^{best}$ and $\widetilde{Y}^{best}$, respectively. We then take $\widehat{Y}^{v/r}_i$ and $\widehat{Y}^{best}$ after argmax to get the difference set between their union and intersection, which is denoted as $\mathcal{M}^{v/r}_{diff}$:
\begin{equation}
\begin{aligned}
\mathcal{M}^{v/r}_{diff} &= \left( \arg\max \widehat{Y}^{v/r}_i \cup \arg\max \widehat{Y}^{best} \right) \\
&\quad - \left( \arg\max \widehat{Y}^{v/r}_i \cap \arg\max \widehat{Y}^{best} \right)
\end{aligned}
\end{equation}
where $v$ and $r$ represent VNet students and ResVNet students, respectively. Similarly, we take $\widehat{Y}^{v/r}_i$ and $Y^L$ after argmax to get the difference set between their union and intersection, which is denoted as $\mathcal{M}^{v/r}_{mis}$:
\begin{equation}
\begin{aligned}
\mathcal{M}^{v/r}_{mis} &= \left( \arg\max \widehat{Y}^{v/r}_i \cup \arg\max Y^L \right) 
\\
&\quad - \left( \arg\max \widehat{Y}^{v/r}_i \cap \arg\max Y^L \right)
\end{aligned}
\end{equation}

Afterward, $\mathcal{M}^{v/r}_{diffmis} = \mathcal{M}^{v/r}_{diff} \odot \mathcal{M}^{v/r}_{mis}$. Ultimately, the loss function $\mathcal{L}_{DFS}$ is derived as the Kullback-Leibler divergence between the student network's output $\widehat{Y}^{v/r}_i$ and the ground truth $Y^L$:
\begin{equation}
\mathcal{L}^{v/r}_{DFS} = \mathcal{L}_{KL}(\widehat{Y}^{v/r}_i, Y^L) \odot \mathcal{M}^{v/r}_{diffmis}
\end{equation}

\begin{table*}[t]
    \small
    \centering
    \caption{Ablation results on FDDI+ dataset}
    \label{Ablation results}
    \resizebox{2\columnwidth}{!}{ 
    \begin{tabular}{cc|cccc|cccc}
        \toprule[1pt]
        \multicolumn{2}{c|}{ScansUsed} & \multicolumn{4}{c|}{Components} & \multicolumn{4}{c}{Metrics} \\
        \cline{1-2} \cline{3-4} \cline{5-6} \cline{7-10} 
        Labeled & Unlabeled & $\mathcal{L}_{s} + \mathcal{L}_{U} + \mathcal{L}_{T}$ & $\mathcal{L}_{KL}$ & $\mathcal{M}_{mis}$ & $\mathcal{M}_{div}$ & Dice (\%) ↑ & Jaccard (\%) ↑ & 95HD (voxel) ↓ & ASD (voxel) ↓ \\
        \midrule[0.5pt]
         & & $\checkmark$ & & & & 86.06 & 75.68 & 91.60 & 17.18 \\
         & & $\checkmark$ & $\checkmark$ & & & 87.67 & 78.04 & 91.01 & 12.22 \\
        11 & 66 & $\checkmark$ & $\checkmark$ & $\checkmark$ & & 87.65 & 78.04 & 48.53 & 9.02 \\
         & & $\checkmark$ & $\checkmark$ & & $\checkmark$ & 88.32 & 79.08 & 9.06 & \textbf{8.10} \\
         & & $\checkmark$ & $\checkmark$ & $\checkmark$ & $\checkmark$ & \textbf{88.47} & \textbf{79.33} & \textbf{8.37} & 8.67 \\
        \bottomrule[1pt]
    \end{tabular}
    }
\end{table*}

\begin{table*}[t]
    \centering
    \caption{Comparison results on FDDI+ dataset with 9\% and 14\% labeled data}
    \label{FDDI+ Table}
    \footnotesize  
    \setlength{\tabcolsep}{3.5pt}  
    \renewcommand{\arraystretch}{1.1}  
    \begin{tabular}{l|cc|cccc}
        \toprule[1pt]
        \multirow{2}{*}{Method} & \multicolumn{2}{c|}{Scans used} & \multicolumn{4}{c}{Metrics} \\
        \cline{2-7}
        & Labeled & Unlabeled & Dice$\uparrow$ & Jaccard$\uparrow$ & 95HD$\downarrow$ & ASD$\downarrow$ \\
        \midrule[0.5pt]
        V-Net\cite{Milletari2016VNet}(3DV2016) & 7 & 0 & 79.88 & 66.67 & 55.30 & 10.10 \\
        ResV-Net\cite{wang2023mcf}(CVPR2023) & 7 & 0 & 80.54 & 67.54 & 53.03 & 9.00 \\
        V-Net\cite{Milletari2016VNet} (3DV2016)& 11 & 0 & 81.93 & 69.47 & 47.24 & 5.12 \\
        ResV-Net\cite{wang2023mcf}(CVPR2023) & 11 & 0 & 81.86 & 69.38 & 50.64 & 6.58 \\
        \midrule[0.5pt]
        UA-MT\cite{yu2019uamt} (MICCAI2019) & \multirow{9}{*}{7(9\%)} & \multirow{9}{*}{70} & 74.60 & 59.55 & 125.44 & 41.87 \\
        SASSNet\cite{li2020sassnet} (MICCAI2020) & & & 77.86 & 63.93 & 145.55 & 50.24 \\
        DTC\cite{luo2021dtc} (AAAI2021) & & & 37.62 & 26.25 & 98.37 & 35.23 \\
        MC-Net+\cite{wu2021mcnet} (MICCAI2021) & & & 75.92 & 61.66 & 136.86 & 40.13 \\
        BCP\cite{bai2023bcp} (CVPR2023) & & & 19.86 & 14.14 & 122.67 & 68.83 \\
        TTMC\cite{chen2024ttmc} (CBM2024) & & & 77.63 & 63.63 & 133.6 & 35.64 \\
        PMT\cite{gao2024pmt} (ECCV2024) & & & \underline{84.91} & \underline{73.85} & \underline{9.85} & \underline{3.57} \\
        SDCL\cite{song2024SDCL} (MICCAI2024) & & & 72.06 & 56.46 & 167.24 & 60.11 \\
        \textbf{RAIL (Ours)} & & & $\textbf{89.55}_{\textcolor{red}{\textbf{↑4.64}}}$ & $\textbf{81.21}_{\textcolor{red}{\textbf{↑7.36}}}$ & $\textbf{6.74}_{\textcolor{red}{\textbf{↓3.11}}}$ & $\textbf{3.20}_{\textcolor{red}{\textbf{↓0.37}}}$ \\
        \midrule[0.5pt]
        UA-MT\cite{yu2019uamt} (MICCAI2019) & \multirow{9}{*}{11(14\%)} & \multirow{9}{*}{66} & 75.39 & 60.82 & 129.82 & 47.19 \\
        SASSNet\cite{li2020sassnet} (MICCAI2020) & & & 77.32 & 63.35 & 135.71 & 42.06 \\
        DTC\cite{luo2021dtc} (AAAI2021) & & & 38.37 & 26.65 & 96.25 & 36.14 \\
        MC-Net+\cite{wu2021mcnet} (MICCAI2021) & & & 81.44 & 68.91 & 135.93 & 33.25 \\
        BCP\cite{bai2023bcp} (CVPR2023) & & & 39.32 & 27.88 & 93.56 & 34.25 \\
        TTMC\cite{chen2024ttmc} (CBM2024) & & & 85.10 & 74.15 & 94.07 & 16.84 \\
        PMT\cite{gao2024pmt} (ECCV2024) & & & \underline{87.37} & \underline{77.64} & \underline{7.81} & \underline{2.67} \\
        SDCL\cite{song2024SDCL} (MICCAI2024) & & & 66.31 & 49.8 & 172.85 & 63.70 \\
        \textbf{RAIL (Ours)} & & & $\textbf{89.75}_{\textcolor{red}{\textbf{↑2.38}}}$ & $\textbf{81.51}_{\textcolor{red}{\textbf{↑3.87}}}$ & $\textbf{6.24}_{\textcolor{red}{\textbf{↓1.57}}}$ & $\textbf{2.32}_{\textcolor{red}{\textbf{↓0.35}}}$ \\
        \bottomrule[1pt]
    \end{tabular}
\end{table*}

\begin{table*}[h]
    \centering
    \caption{Comparison results on FDDI-E dataset with 10\% and 20\% labeled data}
    \label{FDDI-E Table}
    \footnotesize  
    \setlength{\tabcolsep}{4pt}  
    \renewcommand{\arraystretch}{1.1}  
    \begin{tabular}{l|cc|cccc}
        \toprule[1pt]
        \multirow{2}{*}{Method} & \multicolumn{2}{c|}{Scans used} & \multicolumn{4}{c}{Metrics} \\
        \cline{2-7}
        & Labeled & Unlabeled & Dice$\uparrow$ & Jaccard$\uparrow$ & 95HD$\downarrow$ & ASD$\downarrow$ \\
        \midrule[0.5pt]
        V-Net\cite{Milletari2016VNet}(3DV2016) & 20 & 0 & 87.44 & 77.93 & 25.77 & 5.46 \\
        ResV-Net\cite{wang2023mcf}(CVPR2023) & 20 & 0 & 78.94 & 66.86 & 66.10 & 23.90 \\
        V-Net\cite{Milletari2016VNet}(3DV2016) & 40 & 0 & 87.30 & 77.83 & 32.47 & 7.85 \\
        ResV-Net\cite{wang2023mcf}(CVPR2023) & 40 & 0 & 77.45 & 65.11 & 71.86 & 27.30 \\
        V-Net\cite{Milletari2016VNet}(3DV2016) & 200 & 0 & 90.10 & 82.19 & 21.21 & 3.87 \\
        ResV-Net\cite{wang2023mcf}(CVPR2023) & 200 & 0 & 81.01 & 69.72 & 65.80 & 25.00 \\
        \midrule[0.5pt]
        UA-MT\cite{yu2019uamt} (MICCAI2019) & \multirow{9}{*}{20(10\%)} & \multirow{9}{*}{180} & 85.50 & 75.48 & 62.61 & 23.82 \\
        SASSNet\cite{li2020sassnet} (MICCAI2020) & & & 88.56 & 79.78 & 49.07 & 11.55 \\
        DTC\cite{luo2021dtc} (AAAI2021) & & & 71.61 & 58.72 & 43.18 & 1.18 \\
        MC-Net+\cite{wu2021mcnet} (MICCAI2021) & & & 88.06 & 79.15 & 52.63 & 13.25 \\
        BCP\cite{bai2023bcp} (CVPR2023) & & & 71.10 & 58.99 & 42.16 & 4.19 \\
        TTMC\cite{chen2024ttmc} (CBM2024) & & & \underline{89.52} & \underline{81.21} & \underline{41.00} & \underline{8.40} \\
        PMT\cite{gao2024pmt} (ECCV2024) & & & 88.01 & 78.83 & 11.23 & 2.27 \\
        SDCL\cite{song2024SDCL} (MICCAI2024) & & & 86.59 & 76.85 & 60.22 & 17.86 \\
        \textbf{RAIL (Ours)} & & & 
        $\textbf{90.74}_{\textcolor{red}{\textbf{↑1.22}}}$ &
        $\textbf{83.17}_{\textcolor{red}{\textbf{↑1.96}}}$ &
        $\textbf{5.27}_{\textcolor{red}{\textbf{↓35.73}}}$ &
        $\textbf{1.08}_{\textcolor{red}{\textbf{↓7.32}}}$ \\
        \midrule[0.5pt]
        UA-MT\cite{yu2019uamt} (MICCAI2019) & \multirow{9}{*}{40(20\%)} & \multirow{9}{*}{160} & 86.44 & 76.77 & 58.97 & 20.50 \\
        SASSNet\cite{li2020sassnet} (MICCAI2020) & & & \underline{90.60} & \underline{82.98} & \underline{29.38} & \underline{7.32} \\
        DTC\cite{luo2021dtc} (AAAI2021) & & & 71.96 & 59.92 & 43.11 & 2.47 \\
        MC-Net+\cite{wu2021mcnet} (MICCAI2021) & & & 87.42 & 78.36 & 59.01 & 20.16 \\
        BCP\cite{bai2023bcp} (CVPR2023) & & & 66.07 & 55.73 & 53.55 & 14.91 \\
        TTMC\cite{chen2024ttmc} (CBM2024) & & & 90.01 & 83.03 & 24.69 & 5.71 \\
        PMT\cite{gao2024pmt} (ECCV2024) & & & 88.64 & 79.86 & 14.15 & 2.92 \\
        SDCL\cite{song2024SDCL} (MICCAI2024) & & & 85.79 & 75.94 & 65.95 & 22.75 \\
        \textbf{RAIL (Ours)} & & & 
        $\textbf{90.92}_{\textcolor{red}{\textbf{↑0.32}}}$ &
        $\textbf{83.47}_{\textcolor{red}{\textbf{↑0.49}}}$ &
        $\textbf{5.58}_{\textcolor{red}{\textbf{↓23.80}}}$ &
        $\textbf{1.05}_{\textcolor{red}{\textbf{↓6.27}}}$ \\
        \bottomrule[1pt]
    \end{tabular}
\end{table*}

\subsection{Confidence-Aware Learning}

In the context of unsupervised learning, we introduce the Confidence-Aware Learning (CAL) Modulator, which seeks to maximize the uncertainty in regions of divergence between the student network's pseudo-labels and those of the current best-performing student. This strategy mitigates the influence of low-confidence prediction regions during model training, thereby enhancing the stability and reliability of the generated pseudo-labels. The workflow of the Confidence-Aware Learning (CAL) Modulator can be summarized in Algorithm~\ref{CAL Algorithm}.

\captionsetup[figure]{labelfont=bf, labelsep=period, name=Fig.}  

\begin{figure*}[t]
  \centering
  \includegraphics[width=1\textwidth]{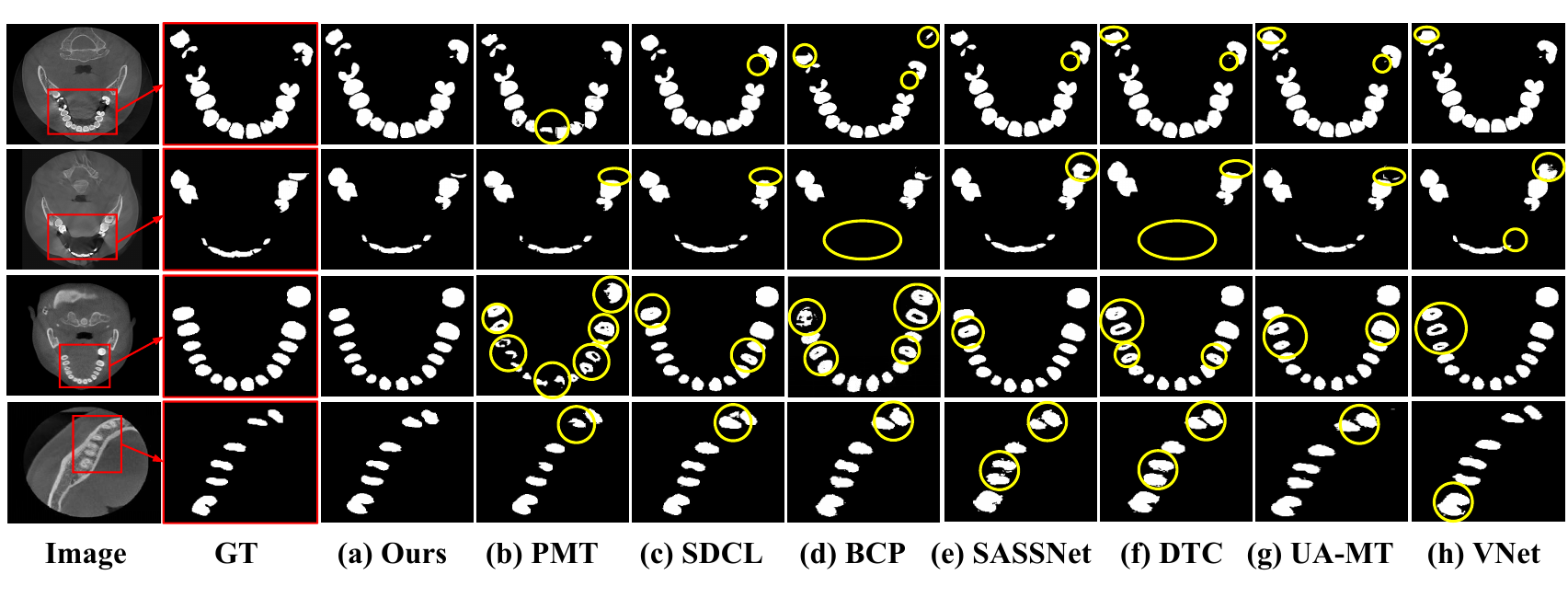}
  \caption{2D segmentation visualization of different semi-supervised methods on FDDI+ (first line), FDDI-E (second line), 3D CBCT Tooth (third line) and CTooth (last line) dataset under 14\%, 10\%, 10\% and 10\% labeled, respectively.}
    \label{fig:2d_vis}
\end{figure*}

\begin{figure*}[t]
  \centering
  \includegraphics[width=1\textwidth]{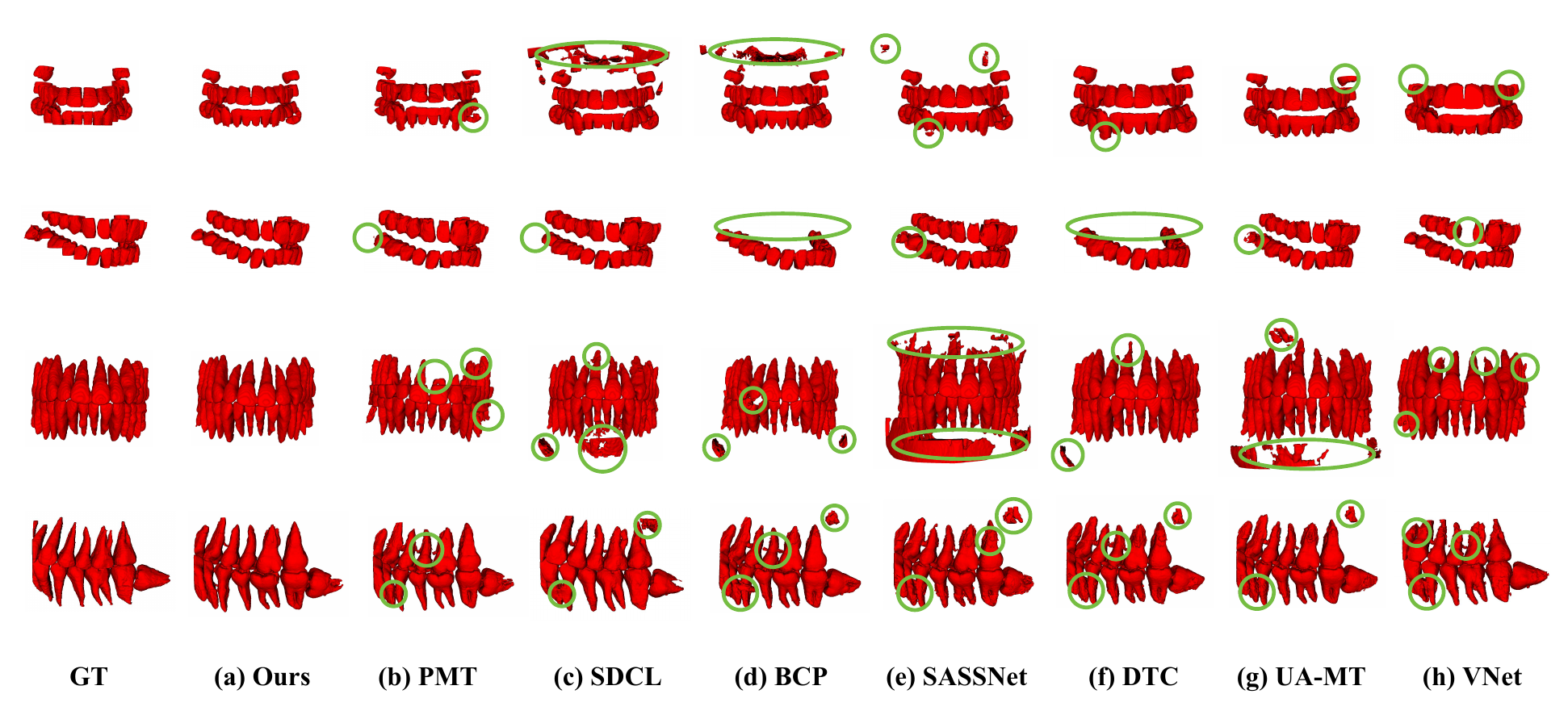}
  \caption{3D segmentation visualization of different semi-supervised methods on FDDI+ (first line), FDDI-E (second line), 3D CBCT Tooth (third line) and CTooth (last line) dataset under 14\%, 10\%, 10\% and 10\% labeled, respectively.}
  \label{fig:3d_vis}
\end{figure*}

Finally, we linearly combine $\mathcal{L}_s$, $\mathcal{L}_{DFS}$, $\mathcal{L}_U$, $\mathcal{L}_T$, $\mathcal{L}_{CAL}$ with specific weights to form the total loss function:
\begin{equation}
\mathcal{L}^{v/r}_{total} = \mathcal{L}^{v/r}_s + \gamma \mathcal{L}^{v/r}_{DFS} + \lambda_1 (\mathcal{L}^{v/r}_U + \mu \mathcal{L}^{v/r}_{CAL}) + \lambda_2 \mathcal{L}^{v/r}_T
\end{equation}
where $\gamma = 0.05$ and $\mu = 0.1$. 

As the training progresses, the values of $\lambda_1$ and $\lambda_2$ increase according to the iteration, reaching a plateau after a certain number of iterations. The PMT method utilizes two independent Gaussian warm-up functions to regulate the weights of the loss functions, $\lambda_1$ and $\lambda_2$, each governed by distinct parameters:
\begin{equation}
\begin{aligned}
\lambda_1(t_i) &= 
\begin{cases}
\hat{\lambda}_1 \cdot e^{-5\left( 1 - \frac{2t_i}{t_{max}} \right)^2}, & t_i < \frac{t_{max}}{2} \\
\hat{\lambda}_1, & t_i \geq \frac{t_{max}}{2}
\end{cases} \\
\lambda_2(t_i) &= 
\begin{cases}
\hat{\lambda}_2 \cdot e^{-5\left( 1 - \frac{2t_i}{t_{max}} \right)^2}, & t_i < \frac{t_{max}}{2} \\
\hat{\lambda}_2, & t_i \geq \frac{t_{max}}{2}
\end{cases}
\end{aligned}
\end{equation}
Here, $t_i$ and $t_{max}$ indicate the current and total training steps. The coefficients $\hat{\lambda}_1$ and $\hat{\lambda}_2$ are empirically initialized to 20.0 and 10.0.

\section{Experiments}
\label{sec:experiments}

\subsection{Datasets and Metrics}

Our method is evaluated on four datasets: FDDI+~\cite{yu2023benchmark}, FDDI-E, 3D CBCT Tooth~\cite{cui2022fully}, and CTooth~\cite{cui2022ctooth}. For each dataset, the training volumes are randomly cropped to a size of 112 × 112 × 80 to serve as model input. To cope with limited GPU memory and sparse labels, 15 patches are extracted from each scan. The cropped volumes are normalized to reduce scanning-induced noise and artifacts before model input. At inference, predictions are generated using a fixed-size sliding window with a stride of 64 × 64 × 32.

\subsubsection{FDDI+ Dataset}

\begin{table*}[htbp]
    \centering
    \caption{Comparison results on 3D CBCT Tooth dataset with 5\% and 10\% labeled data}
    \label{3D CBCT Table}
    \footnotesize  
    \setlength{\tabcolsep}{4pt}  
    \renewcommand{\arraystretch}{1.1}  
    \begin{tabular}{l|cc|cccc}
        \toprule[1pt]
        \multirow{2}{*}{Method} & \multicolumn{2}{c|}{Scans used} & \multicolumn{4}{c}{Metrics} \\
        \cline{2-7}
        & Labeled & Unlabeled & Dice$\uparrow$ & Jaccard$\uparrow$ & 95HD$\downarrow$ & ASD$\downarrow$ \\
        \midrule[0.5pt]
        V-Net\cite{Milletari2016VNet}(3DV2016) & 7 & 0 & 89.39 & 81.33 & 2.75 & 0.85 \\
        ResV-Net\cite{wang2023mcf}(CVPR2023) & 7 & 0 & 79.34 & 65.90 & 19.25 & 5.17 \\
        V-Net\cite{Milletari2016VNet}(3DV2016) & 13 & 0 & 93.51 & 87.98 & 1.48 & 0.52 \\
        ResV-Net\cite{wang2023mcf}(CVPR2023) & 13 & 0 & 92.61 & 86.40 & 1.65 & 0.60 \\
        V-Net\cite{Milletari2016VNet}(3DV2016) & 120 & 0 & 94.55 & 89.70 & 1.24 & 0.40 \\
        ResV-Net\cite{wang2023mcf}(CVPR2023) & 120 & 0 & 92.99 & 86.94 & 1.48 & 0.75 \\
        \midrule[0.5pt]
        UA-MT\cite{yu2019uamt} (MICCAI2019) & \multirow{8}{*}{7(5\%)} & \multirow{8}{*}{113} & 85.83 & 75.33 & 25.70 & 5.68 \\
        SASSNet\cite{li2020sassnet} (MICCAI2020) & & & 85.16 & 74.25 & 33.45 & 6.91 \\
        DTC\cite{luo2021dtc} (AAAI2021) & & & \underline{90.06} & \underline{82.19} & \underline{4.96} & \underline{1.76} \\
        MC-Net+\cite{wu2021mcnet} (MICCAI2021) & & & 88.38 & 79.39 & 16.91 & 3.30 \\
        BCP\cite{bai2023bcp} (CVPR2023) & & & 84.29 & 74.04 & 17.43 & 0.64 \\
        TTMC\cite{chen2024ttmc} (CBM2024) & & & 80.34 & 68.22 & 16.00 & 0.69 \\
        PMT\cite{gao2024pmt} (ECCV2024) & & & 85.00 & 74.83 & 3.57 & 1.45 \\
        SDCL\cite{song2024SDCL} (MICCAI2024) & & & 86.05 & 75.72 & 30.27 & 6.01 \\
        \textbf{RAIL (Ours)} & & & 
        $\textbf{91.60}_{\textcolor{red}{\textbf{↑1.54}}}$ &
        $\textbf{84.73}_{\textcolor{red}{\textbf{↑2.54}}}$ &
        $\textbf{2.03}_{\textcolor{red}{\textbf{↓2.93}}}$ &
        $\textbf{0.62}_{\textcolor{red}{\textbf{↓1.14}}}$ \\
        \midrule[0.5pt]
        UA-MT\cite{yu2019uamt} (MICCAI2019) & \multirow{9}{*}{13(10\%)} & \multirow{9}{*}{107} & 91.06 & 83.81 & 16.55 & 3.57 \\
        SASSNet\cite{li2020sassnet} (MICCAI2020) & & & 81.10 & 68.73 & 28.15 & 7.82 \\
        DTC\cite{luo2021dtc} (AAAI2021) & & & \underline{92.68} & \underline{86.48} & \underline{2.04} & \underline{1.49} \\
        MC-Net+\cite{wu2021mcnet} (MICCAI2021) & & & 92.01 & 85.42 & 3.14 & 2.00 \\
        BCP\cite{bai2023bcp} (CVPR2023) & & & 87.17 & 77.96 & 9.67 & 0.52 \\
        TTMC\cite{chen2024ttmc} (CBM2024) & & & 76.15 & 64.61 & 43.15 & 0.55 \\
        PMT\cite{gao2024pmt} (ECCV2024) & & & 87.93 & 79.10 & 2.75 & 0.82 \\
        SDCL\cite{song2024SDCL} (MICCAI2024) & & & 90.72 & 83.15 & 14.29 & 3.42 \\
        \textbf{RAIL (Ours)} & & & 
        $\textbf{94.09}_{\textcolor{red}{\textbf{↑1.41}}}$ &
        $\textbf{88.96}_{\textcolor{red}{\textbf{↑2.48}}}$ &
        $\textbf{1.31}_{\textcolor{red}{\textbf{↓0.73}}}$ &
        $\textbf{0.44}_{\textcolor{red}{\textbf{↓1.05}}}$ \\
        \bottomrule[1pt]
    \end{tabular}
\end{table*}

This study primarily utilizes the Fudan Dual-Modality Dental Imaging (FDDI) dataset~\cite{yu2023benchmark}, which consists of 66 CBCT scans. Additionally, we collect 14 supplementary scans to enhance our experimental analysis, termed as FDDI+ dataset. Informed consent is obtained from all patients, and all original DICOM images are anonymized to ensure privacy. Each scan, acquired using clinical-grade medical imaging instruments, comprises 400 axial slices with a resolution of \(800 \times 800\) with 1mm slice thickness. 

To comprehensively evaluate the proposed method, we utilize a total of 80 3D CBCT scans and design two experimental settings. In the first setting, training includes 7 labeled (9\%) and 70 unlabeled scans, while 3 scans are reserved for testing. The second setting adopts 11 labeled (14\%) and 66 unlabeled scans for training, with 3 held out for evaluation.

\subsubsection{FDDI-E Dataset}

The second dataset employed in this research is the FDDI-E dataset, an extended version of the original FDDI dataset. The FDDI-E dataset contains 286 CBCT scans and the corresponding labels, and their dimensional size is \(604 \times 604 \times 412\). During the experiments, we designed two experimental settings. In the first experimental configuration, 20 labeled volumes (10\%) and 180 unlabeled volumes constitute the training set, while 30 labeled volumes are reserved for validation and 56 labeled volumes for testing. In the second configuration, the training set comprises 40 labeled volumes (20\%) along with 160 unlabeled volumes, maintaining the same validation and test partitions of 30 and 56 labeled volumes, respectively.

\subsubsection{3D CBCT Tooth Dataset}

A subset of the CBCT dataset from Cui et al.~\cite{cui2022fully} is used, comprising 4,938 CBCT scans obtained from 15 medical centers across China, representing a wide range of data distributions. Due to privacy and regulatory restrictions, only a portion of this dataset is publicly available. For our experiments, we utilize 126 3D CBCT scans and implement two experimental configurations to assess the proposed method. In the first configuration, 7 labeled scans (5\%) and 113 unlabeled scans are used for training, with 6 labeled scans reserved for testing. The second configuration uses 13 labeled (10\%) and 107 unlabeled samples for training, with 6 labeled for evaluation.

\subsubsection{CTooth Dataset}

The CTooth dataset~\cite{cui2022ctooth} includes a total of 131 scans, with 22 labeled and 109 unlabeled, providing a comprehensive resource for segmentation tasks. To evaluate our method, we design two experimental settings using 122 scans for training and 7 for testing. In the first setting, there are 7 labeled (5\%) and 115 unlabeled for training. In the second setting, there are 13 labeled (10\%) and 109 unlabeled for training.

\subsubsection{Metrics}

In line with previous works \cite{bai2023bcp,li2020sassnet}, \cite{luo2021dtc}, \cite{wang2023mcf}, \cite{xu2022all}, \cite{yu2019uamt}, we evaluate model performance using four key metrics. These include regional sensitivity measures, such as the Dice similarity coefficient (Dice) \cite{yu2019uamt} and the Jaccard similarity coefficient (Jaccard) \cite{luo2021dtc}, as well as edge-sensitive metrics, including the 95\% Hausdorff Distance (95HD) \cite{xu2022all} and the Average Surface Distance
(ASD) \cite{bai2023bcp}.

\subsection{Implementation Details}

All experiments were run on an NVIDIA RTX 4090 24GB using Ubuntu 20.04 and PyTorch 1.11.0. We employ PMT~\cite{gao2024pmt}, a Mean Teacher-based semi-supervised baseline. The final prediction aggregates outputs from four student models. Training uses SGD (momentum = 0.9, weight decay = 0.0004) with an initial learning rate of 0.01 and linear warm-up over the first 1,000 iterations. After reaching 4,000 iterations, it is progressively reduced to 1e-5 following a cosine annealing schedule~\cite{loshchilov2016sgdr}, with a total of 8,000 training iterations. The batch size of 2 is employed, where each batch comprises a single labeled sample alongside an unlabeled one. The hyperparameters are configured as $\alpha = 0.5$, $\beta = 0.05$. 

\subsection{Ablation Study}
Table~\ref{Ablation results} presents an ablation analysis evaluating the individual and combined contributions of key components within our framework, based on Dice score improvements over the baseline. The results demonstrate that both the Disagreement-Focused Supervision (DFS) Controller and the Confidence-Aware Learning (CAL) Modulator in the RAIL architecture contribute positively to segmentation performance. Importantly, the highest performance is achieved when both $\mathcal{M}_{mis}$ and $\mathcal{M}_{div}$ are jointly applied, underscoring their synergistic effect.

\subsection{Compare with Other Methods}

\begin{table*}[h]
    \centering
    \caption{Comparison results on CTooth dataset with 5\% and 10\% labeled data}
    \label{CTooth Table}
    \footnotesize  
    \setlength{\tabcolsep}{4pt}  
    \renewcommand{\arraystretch}{1.1}  
    \begin{tabular}{l|cc|cccc}
        \toprule[1pt]
        \multirow{2}{*}{Method} & \multicolumn{2}{c|}{Scans used} & \multicolumn{4}{c}{Metrics} \\
        \cline{2-7}
        & Labeled & Unlabeled & Dice$\uparrow$ & Jaccard$\uparrow$ & 95HD$\downarrow$ & ASD$\downarrow$ \\
        \midrule[0.5pt]
        V-Net\cite{Milletari2016VNet}(3DV2016) & 7 & 0 & 88.09 & 78.98 & 7.01 & 1.50 \\
        ResV-Net\cite{wang2023mcf}(CVPR2023) & 7 & 0 & 87.06 & 77.31 & 8.06 & 2.10 \\
        V-Net\cite{Milletari2016VNet}(3DV2016) & 13 & 0 & 88.34 & 79.44 & 7.11 & 1.51 \\
        ResV-Net\cite{wang2023mcf}(CVPR2023) & 13 & 0 & 87.48 & 77.99 & 6.90 & 1.68 \\
        \midrule[0.5pt]
        UA-MT\cite{yu2019uamt} (MICCAI2019) & \multirow{9}{*}{7(5\%)} & \multirow{9}{*}{115} & 86.04 & 75.67 & 11.54 & 3.70 \\
        SASSNet\cite{li2020sassnet} (MICCAI2020) & & & 81.38 & 68.81 & 29.48 & 7.30 \\
        DTC\cite{luo2021dtc} (AAAI2021) & & & 84.90 & 73.97 & 11.6 & 3.81 \\
        MC-Net+\cite{wu2021mcnet} (MICCAI2021) & & & 87.31 & 77.60 & 6.74 & 2.29 \\
        BCP\cite{bai2023bcp} (CVPR2023) & & & 81.01 & 70.09 & 32.53 & 1.84 \\
        TTMC\cite{chen2024ttmc} (CBM2024) & & & 79.50 & 67.59 & 25.81 & 3.04 \\
        PMT\cite{gao2024pmt} (ECCV2024) & & & \underline{88.09} & \underline{78.83} & \underline{6.49} & \underline{1.58} \\
        SDCL\cite{song2024SDCL} (MICCAI2024) & & & 85.13 & 74.53 & 15.68 & 5.07 \\
        \textbf{RAIL (Ours)} & & & 
        $\textbf{89.36}_{\textcolor{red}{\textbf{↑1.27}}}$ &
        $\textbf{81.01}_{\textcolor{red}{\textbf{↑2.18}}}$ &
        $\textbf{5.48}_{\textcolor{red}{\textbf{↓1.01}}}$ &
        $\textbf{1.45}_{\textcolor{red}{\textbf{↓0.13}}}$ \\
        \midrule[0.5pt]
        UA-MT\cite{yu2019uamt} (MICCAI2019) & \multirow{9}{*}{13(10\%)} & \multirow{9}{*}{109} & 86.57 & 76.54 & 10.52 & 3.52 \\
        SASSNet\cite{li2020sassnet} (MICCAI2020) & & & 85.63 & 75.10 & 9.76 & 3.48 \\
        DTC\cite{luo2021dtc} (AAAI2021) & & & 85.03 & 74.15 & 10.58 & 3.32 \\
        MC-Net+\cite{wu2021mcnet} (MICCAI2021) & & & 84.72 & 73.70 & 9.41 & 3.45 \\
        BCP\cite{bai2023bcp} (CVPR2023) & & & 80.12 & 68.42 & 33.67 & 2.18 \\
        TTMC\cite{chen2024ttmc} (CBM2024) & & & 80.86 & 68.56 & 20.59 & 4.88 \\
        PMT\cite{gao2024pmt} (ECCV2024) & & & 86.74 & 76.82 & 7.90 & 2.44 \\
        SDCL\cite{song2024SDCL} (MICCAI2024) & & & \underline{88.43} & \underline{79.51} & \underline{8.46} & \underline{3.41} \\
        \textbf{RAIL (Ours)} & & & 
        $\textbf{89.03}_{\textcolor{red}{\textbf{↑0.6}}}$ &
        $\textbf{80.49}_{\textcolor{red}{\textbf{↑0.98}}}$ &
        $\textbf{6.03}_{\textcolor{red}{\textbf{↓2.43}}}$ &
        $\textbf{1.58}_{\textcolor{red}{\textbf{↓1.83}}}$ \\
        \bottomrule[1pt]
    \end{tabular}
\end{table*}

We conduct a comprehensive comparison between our method and existing SOTA approaches across four datasets: FDDI+ dataset, FDDI-E dataset, 3D CBCT Tooth dataset, and CTooth dataset. 

PMT serves as the primary baseline for evaluation. In addition, we benchmark against several representative methods, including UA-MT~\cite{yu2019uamt}, which introduces uncertainty-aware self-ensembling; SASSNet~\cite{li2020sassnet}, which incorporates geometric shape constraints; and DTC~\cite{luo2021dtc}, which exploits dual-task consistency for enhanced structural prediction. We also include MC-Net+~\cite{wu2021mcnet} with dual-decoder mutual consistency, TTMC~\cite{chen2024ttmc}, introducing a triple-task mutual consistency framework, and BCP~\cite{bai2023bcp}, which employs bidirectional Copy-Paste to align labeled and unlabeled data distributions. Additionally, SDCL~\cite{song2024SDCL} enhances semi-supervised segmentation by incorporating student discrepancy-informed correction learning. 

For fair comparison, all methods are configured according to their official settings, with training capped at 8,000 iterations. BCP and SDCL are pre-trained for 2,000 iterations and fine-tuned for the remaining 6,000.

\subsubsection{Comparison on FDDI+ Dataset}

We evaluate our model on FDDI+ under 9\% and 14\% label ratios. As shown in Table~\ref{FDDI+ Table}, it consistently outperforms PMT and recent strong baselines across all four metrics. 

With only 9\% labeled data, our approach surpasses the strongest competing method by margins of 4.64\% in Dice and 7.36\% in Jaccard, while reducing 95HD and ASD by 3.11 and 0.37, respectively. Under the 14\% setting, our model continues to deliver superior performance, yielding gains of 2.38\% in Dice and 3.87\% in Jaccard, along with reductions of 1.57 in 95HD and 0.35 in ASD. Remarkably, even with a smaller fraction of labeled samples, our framework outperforms PMT trained on 14\% labeled data, demonstrating enhanced label efficiency and generalization capability.  

To further illustrate the effectiveness of our approach, we provide 2D and 3D qualitative visualizations of segmentation results in Fig.~\ref{fig:2d_vis} and Fig.~\ref{fig:3d_vis}, respectively. The visual comparisons emphasize the superior segmentation quality of our method across different proportions of labeled data, demonstrating its robustness in tackling the challenging FDDI+ dataset.
\subsubsection{Comparison on FDDI-E Dataset}

We evaluate performance under 10\% and 20\% labeling protocols. As listed in Table~\ref{FDDI-E Table}, our method consistently yields superior results over PMT and recent semi-supervised techniques across all four metrics.

When trained with 10\% labeled data, our approach delivers performance gains of 1.22\% in Dice and 1.96\% in Jaccard, along with substantial reductions of 35.73 in 95HD and 7.32 in ASD. Under the 20\% labeling scenario, our model continues to outperform the baseline, yielding an additional 0.32\% improvement in Dice, 0.49\% in Jaccard, a decrease of 23.80 in 95HD, and a 6.27 reduction in ASD.

Fig.~\ref{fig:2d_vis} and Fig.~\ref{fig:3d_vis} illustrate representative 2D and 3D segmentation results on the FDDI-E dataset, offering a clear visual perspective on model performance. As shown, our approach consistently delivers more accurate and refined segmentation across various labeling ratios.

\subsubsection{Comparison on 3D CBCT Tooth Dataset}

Table~\ref{3D CBCT Table} provides a detailed comparison between our framework and previous leading methods, along with the full supervision bounds. The evaluation is conducted under two annotation ratios (5\% and 10\%), and visual results in both 2D and 3D (Fig.~\ref{fig:2d_vis} and Fig.~\ref{fig:3d_vis}) further demonstrate the effectiveness of our framework in segmenting the 3D CBCT Tooth dataset under varying levels of annotation. 

Across both labeling scenarios, our model consistently outperforms existing methods across all four evaluation metrics. Specifically, under the 5\% labeled setting, it achieves improvements of 1.54\% in Dice and 2.54\% in Jaccard, with corresponding reductions of 2.93 and 1.14 in 95HD and ASD, respectively. When the label ratio is increased to 10\%, the model maintains its advantage, yielding gains of 1.41\% in Dice and 2.49\% in Jaccard, while decreasing 95HD and ASD by 0.73 and 1.05, respectively. 

Overall, the results underscore our model's resilience across varying label proportions, consistently outperforming earlier methods on the densely annotated 3D CBCT Tooth dataset.

\subsubsection{Comparison on CTooth Dataset}

We evaluate our model on the CTooth dataset under 5\% and 10\% labeling ratios. Table~\ref{CTooth Table}shows that our method surpasses PMT and other SOTA baselines in all four metrics. Under the 5\% supervision setting, our model achieves gains of 1.27\% in Dice and 2.18\% in Jaccard, along with reductions of 1.01 in 95HD and 0.13 in ASD, when compared to the strongest competing method. At the 10\% annotation level, further improvements are observed, including 1.54\% and 2.48\% increases in Dice and Jaccard, respectively, and decreases of 0.73 in 95HD and 1.05 in ASD.  

Interestingly, the results suggest that the inclusion of additional labeled data does not yield substantial performance gains on this dataset, likely due to the suboptimal annotation quality of the CTooth dataset. This observation is further supported by the fully supervised VNet model, which shows minimal improvement between the two settings. In contrast, our method consistently achieves SOTA performance across both label proportions, indicating its robustness in handling datasets with noisy annotations. These findings highlight the effectiveness of our approach, even in scenarios where annotation quality is a limiting factor.  

To facilitate a more intuitive understanding of model performance on the CTooth dataset, we present representative 2D and 3D qualitative results in Fig.\ref{fig:2d_vis} and Fig.\ref{fig:3d_vis}, respectively. These visual comparisons further confirm the superiority of our method over existing approaches across varying annotation ratios in the context of the complex CTooth segmentation task.
\section{Conclusion}
In this paper, we propose Region-Aware Instructive Learning (RAIL), a novel dual-group, dual-student semi-supervised framework designed for 3D tooth segmentation from CBCT scans. The RAIL model incorporates several innovative mechanisms, including a dual-group, dual-student Mean Teacher framework, the Disagreement-Focused Supervision (DFS) Controller, and the Confidence-Aware Learning (CAL) Modulator. The dual-group, dual-student framework allows for alternating training between two student models, fostering inter-group knowledge transfer and reducing overfitting. The DFS Controller specifically targets regions with structural ambiguity or incorrect labeling, guiding the model to focus on challenging areas and significantly improving prediction accuracy in those regions. Meanwhile, the CAL Modulator adjusts the model’s attention to regions of low confidence, thus stabilizing the learning process by minimizing the impact of unreliable pseudo-labels, which enhances the model’s robustness. Additionally, we conduct extensive experiments with existing state-of-the-art semi-supervised methods, showing that RAIL consistently outperforms other approaches across several benchmark datasets, including FDDI+, FDDI-E, 3D CBCT Tooth, and CTooth. RAIL not only achieves superior segmentation accuracy but also exhibits greater robustness when trained with limited labeled data.

In future work, we aim to enhance the efficiency of the RAIL framework, explore its application to additional modalities, and further improve its ability to generalize to other medical image segmentation tasks. 
{
    \small
    \bibliographystyle{ieeenat_fullname}
    \bibliography{main}
}

\end{document}